\documentclass[11pt,a4paper]{article}
\usepackage[hyperref]{emnlp2020}
\usepackage{times}
\usepackage{latexsym}
\usepackage{graphicx}
\usepackage{makecell}
\usepackage{amsfonts} 
\usepackage{multirow}
\usepackage{lipsum}
\usepackage{amsmath}
\usepackage[justification=centering]{caption}

\usepackage{microtype}

\aclfinalcopy 

\title{Commonsense Knowledge Graph Reasoning by \\ Selection or Generation? Why?}

\author{Cunxiang Wang\textsuperscript{1,2}, Jinhang Wu\textsuperscript{3}, Luxin Liu\textsuperscript{3} and Yue Zhang\textsuperscript{1} \\
\textsuperscript{1}School of Engineering, Westlake University, China\\
\textsuperscript{2}Zhejiang University, China, 
\textsuperscript{3}Alibaba Inc, China\\
  {\tt \{wangcunxiang, zhangyue\}@westlake.edu.cn}\\
  {\tt \{jinhang.wjh, xique.llx\}@alibaba-inc.com}
  }

\date{}

\begin{document}
\maketitle
\begin{abstract}
\vspace{-1mm}
Commonsense knowledge graph reasoning (CKGR) is the task of predicting a missing entity given one existing and the relation in a commonsense knowledge graph (CKG). Existing methods can be classified into two categories \textit{generation method} and \textit{selection method}. 
Each method has its own advantage. 
We theoretically and empirically compare the two methods, finding the \textit{selection method} is more suitable than the \textit{generation method} in CKGR. Given the observation, we further combine the structure of neural Text Encoder and Knowledge Graph Embedding models to solve the \textit{selection method}'s two problems, achieving competitive results. We provide a basic framework and baseline model for subsequent CKGR tasks by selection methods.

\end{abstract}

\section{Introduction}
\vspace{-2mm}

Common sense has received increasing attention in the natural language processing community \citep{ATOMIC, comet}. 
It has been found that current AI systems lack commonsense knowledge {bengio2019from}. 
The \textbf{Commonsense Knowledge Graph} (CKG) \citep{ConceptNet5, ATOMIC}  can be a promising way to introduce structured and explainable commonsense knowledge into AI/NLP systems \citep{CRC, kagnet, senmaking}. Similar to general KGs, such as FreeBase \citep{FreeBase}, CKGs consist of tuples structure as $\langle source\ entity, relation, target\ entity \rangle$. However, compared to general KGs, CKGs have two significant differences. First, CKGs emphasize on commonsense knowledge. Second, entities in CKGs are semantically-rich short text rather than noun entities. 

\begin{figure}[t]
  \center {\includegraphics[width=8cm] {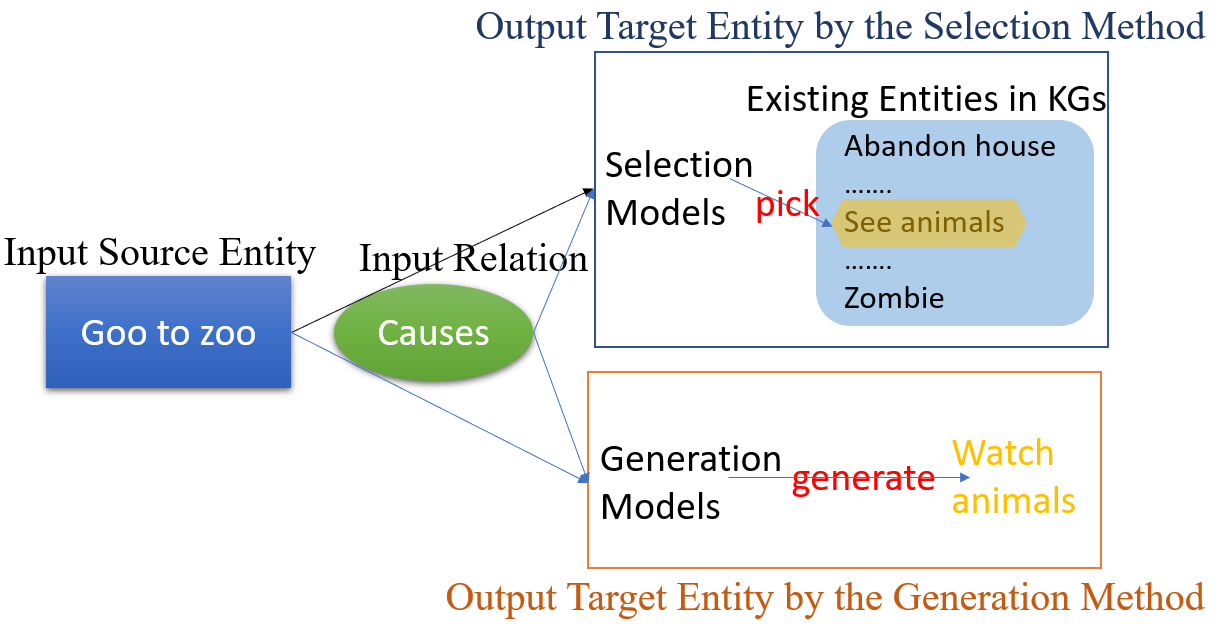}}
  \vspace{-5mm}
  \caption{\label{SelectionGeneration} The comparison between the selection method and generation method on CKGR task.}
\vspace{-6mm}
\end{figure}

CKGs are data-sparse and costly to construct. So, automated methods can be useful to expand the existing CKGs, which we call \textbf{commonsense knowledge graph reasoning} (CKGR). The task is to find one or several correct target entities as output given a source entity and a relation as input. For example, given \textit{``go to zoo"} as source entity and \textit{``Causes"} as relation, we can infer that \textit{``see animal"} is a suitable target entity. Shown in the Figure~\ref{SelectionGeneration}, target entities can either be obtained by selecting from existing entities or machine generation.

Broadly speaking, CKGR methods can be classified into two categories -- the \textit{generation method} \citep{comet} and the \textit{selection method} \citep{transE}. In particular, given one entity and the relation, generation methods synthesize the missing entity as a text sequence. In contrast, selection methods score a list of candidate entities for ranking. The two types of methods have been relatively separately investigated, and there has been little comparison between them.

We find that generation methods have two main disadvantages. The first is the \textit{new semantic entities problem}. In particular, the most important advantage claimed for generation methods is that it can generate \textit{new} entities that do not appear in the dataset. However, when we analyze the generated results of COMET \citep{comet}, we find that there is \textit{no new semantic entities}. Most generated entities can be directly found in training data, and the remaining generating entities also have corresponding training entities with similar meanings. The second is the \textit{human evaluation problem}. It can be difficult to automatically evaluate the \textit{generated} results in commonsense areas \citep{Edunov2019OnTE}, and it is costly and relatively unreproducible.

\begin{table*}[!t]
  \centering
  \setlength{\tabcolsep}{5mm}
  \begin{tabular}{c|cc}
  \hline
  \makecell[c]{\textbf{Dataset}} 
  & \makecell[c]{\textbf{ In the Training set}} 
  & \makecell[c]{\textbf{ Not in the Training set}}\\
  \hline
  \makecell[c]{\textbf{ConceptNet-100k} }& \makecell[c]{1157 (96.42\%)} & \makecell[c]{43 (3.58\%)}\\
  \makecell[c]{\textbf{ATOMIC} } & \makecell[c]{868 (96.44\%)} & 32 (3.56\%)\footnotemark[2] \\
\hline
\multicolumn{3}{c}{(a) How many generated entities are in or not in the training set.} 
\vspace{2mm}
\\ 
\hline
    \makecell[c]{\textbf{Dataset}} 
  &\makecell[c]{\textbf{ Exists similar/better Entity}\\ \textbf{in Training set}}
  &\makecell[c]{\textbf{ No similar/better Enity}\\ \textbf{in the Training set}} \\
\hline

\makecell[c]{\textbf{ConceptNet-100k}} &\makecell[c]{43 (3.58\%)}&\makecell[c]{0 (0.00\%)}\\
\makecell[c]{\textbf{ATOMIC}} &\makecell[c]{32 (3.56\%)}&\makecell[c]{0 (0.00\%)}\\
\hline
\multicolumn{3}{c}{\makecell[c]{(b)  Analysis of the generated entities which do not directly appear in the training set.}} \\
  \end{tabular}
  \vspace{-2mm}
  \caption{\label{genrated-entities-details} Analysis on the generated entities by COMET. We chose the ``greedy decoding" results, which are reported having the highest human evaluation scores \citep{comet}. 
  }
  \vspace{-5mm}
\end{table*}

The selection method was the dominant method for general relational KGs \citep{NTN}. 
Knowledge Graph Embedding (KGE) models, such as TransE \citep{transE} and TuckER \citep{TuckER}, are the most representative selection method models. 
Recently, \citet{malaviya2019exploiting} used the \textit{selection method} on CKGs. 
Compared to the generation method, the selection method has several benefits. 
First, it is relatively easy to evaluate, and we can easily control whether a given tuple structure as $\langle source\ entity, relation, target\ entity \rangle$ is correct or not. 
Second, it can avoid unexpected outputs, such as incorrect or offensive entities, which is beneficial in most industrial applications. 

One issue of selection methods is that they cannot directly deal with the rich textual information in CKGs. In addition, if an entity of a test tuple is \textit{unseen} in the training set, it does not have a trained embedding, which results in the OOV situation where the tuple cannot be scored. 
The \textit{unseen-entities problem} is the crucial difference between CKG \textit{reasoning} and traditional knowledge graph \textit{completion} \citep{transE}. 
All source entities in the test set of the CKG ATOMIC \citep{ATOMIC} are \textit{unseen}. \citet{malaviya2019exploiting} circumvented the unseen problem by re-splitting the test set of ATOMIC, avoiding test entities that are not existent in the training data. However, in practical application scenarios, we cannot expect that all input entities to already exist in the database.

Given the above observation, we combine the architecture of neural text encoders, such as CNN and LSTM, with the standard KGE models, such as TransE \citep{transE} and TuckER \citep{TuckER}. With text sequence node embedding, all KGE models can be simply applied to the CKGR task. 
Experiments on two representative datasets show that our model can achieve competitive results in the CKGR task, showing its effectiveness in helping a \textit{selection method} solve \textit{rich textual information problem} and \textit{unseen-entities problem}. To our knowledge, we are the first to build a selection method for CKGR that allows unseen entities in the test data. \footnotemark[1]

\footnotetext[1]{We will release our data and source code at $ \langle URL \rangle $ (removed for submission).}





\vspace{0mm}


\section{Datasets-CKGs}
\label{ckgs}
We focus on two predominant CKGs, namely ConceptNet (ConceptNet-100K) \citep{KBC} and ATOMIC \citep{ATOMIC}. 
The detailed statistics of the two datasets are listed in Appendix Table 1.

\textbf{ConceptNet-100K} is a sub graph obtained from the Open Mind Common Sense (OMCS) entries in the ConceptNet5 dataset \citep{ConceptNet5} by \citet{KBC}. 
It contains 100,000 tuples as the training set, 1,200 as the dev set, and 1,200 as the test set. 
A tuple in ConceptNet-100K is 
\textit{``go to zoo"  ``causes"  ``see animal"}. The percentage of \textit{Unseen Source Entities} in the test set is 2.8\%.

\textbf{ATOMIC} is a daily event CKG. 
It contains 709,996 tuples as the training set, 79,600 as the dev set, and 87,481 as the test set. The source entities are typically described as ``PersonX does something (on/with/.../ PersonY)", for example, ``PersonX puts his trust in PersonY". 
ATOMIC contains 9 types of relations, including ``xAttribute", ``xIntent", ``xReact", ``xEffect", ``xNeed", ``xWant" and ``oReact", ``oWant", ``oEffect", here ``x" represents ``PersonX" and ``o" represents ``PersonY". The test source entities are \textbf{all} \textit{unseen} at training set.

\section{Analysis of the Generation Method}

There are two representative \textbf{generation methods} for CKGR, including CKB Generation \citep{CKBCG} and COMET \citep{comet}. 
We focus on COMET, which is a ``seq2seq" model that uses a Transformer language model architecture (GPT) \citep{GPT} to generate target entities.

Original experimental results of COMET on ConceptNet and ATOMIC from the authors \citep{comet}, include the human evaluation on 1,200 instances from ConceptNet-100k and 900 instances from ATOMIC. We conduct analysis on the same 1,200 and 900 instances.
The results are shown in Table \ref{genrated-entities-details}. Most generated entities have already appeared in the training set (96.42\% for ConceptNet-100K; 96.44\% for ATOMIC). We manually compared and found that even for the remaining entities (3.58\% for ConceptNet-100K; 3.56\% for ATOMIC), there also exist similar or better entities in the training set. For example, the input source entity is \textit{``read newspaper"}. The input relation is \textit{``motivated by goal"}. COMET's output is \textit{``know about current event"}, which does not appear in the training data, but there is a similar entity \textit{``learn about current event"} in the training data. 

We list all \textit{new} output target entities by COMET, which do not directly appear in the training data and their similar/better counterparts from training data in the Appendix.

\footnotetext[2]{There is a similar metric $N/T_{o}$ in COMET paper, which is 9.71\%. It is because they directly compare characters of two entities, which makes ``being hungry" a different entity from ``Being hungry". We can also get about 9\% results if we use this kind of comparison. But we lowercase all entities first and then compare, which leads to different results.}

\section{Integration Text Encoding to a Selection Model}


Following previous works on KGE \citep{transE, TuckER}, the CKGR task can be described as follows. For any test tuple $\langle s\ r\ t \rangle$, the input is $``s\ r"$, the output is suitable $``t" $s from existing entities. To measure the model performance, for each test tuple $\langle s\ r\ t \rangle$, every existing entity $e_k$ in the CKG will be seen as a candidate target entity and makes a candidate tuple $\langle s\ r\ e_k \rangle $. We calculate every candidate tuple's \textbf{confidence score} and compute the rank ${rk}_t$ of the tuple with the correct $``t"$. Following \citet{TuckER}, we use ${rk}_t$ to measure metrics Mean Rank (MR), Mean Reciprocal Rank (MRR), and HITS@10/3/1. Mean Rank is the mean of ${rk}_t$s, Mean Reciprocal Rank is the mean of the inverse of ${rk}_t$s. Hits@k measures the percentage of ${rk}_t$s within the top k. Lower MR, MRR and higher Hits@10/3/1 indicate the better
results and better performance. Since there might also be other correct tuples with the same $``s\ r"$ but different $``t"$s in training, dev or testing sets, we remove them when computing ${rk}_t$.

\subsection{Model}
We aim to extend standard KGE models for handling text and \textit{unseen entities}. 
The model consists of two modules: 
Text Encoder module and the standard KGE module.
We call the model as \textbf{T}ext \textbf{E}ncoder \textbf{E}nhanced (Encoder Model Name + KGE Model Name), such as \textbf{TEE} (CNN + TuckER). 
For each tuple $\langle s\ r\ t \rangle$. 
\textbf{Text Encoder Module} is used to encode the embeddings of each word of the source entity, relation, and target entity into three embedding vectors. Then the three embedding vectors are used in a \textbf{KGE Module} to calculate the confidence score of the tuple. 

In the experiments, we choose two relatively simple models (i.e., CNN and BiLSTM) as the text encoder models. 
For the KGE Module, we use one representative KGE model TransE \citep{transE} and one state-of-art KGE model TuckER \citep{TuckER}, 
which serve as the baseline models in the experiments. 
In general, other text encoder models and KGE models can also be used.







\begin{table*}[htb]
  \centering
  \small
  \setlength{\tabcolsep}{1mm} 
  \label{label:link_prediction_rel}
  \begin{tabular}{c|ccc|ccc}
  \hline
  \multirow{3}{*}{\textbf{Model}} & \multicolumn{3}{c|}{\textbf{ATOMIC}} & \multicolumn{3}{c}{\textbf{ConceptNet-100k}} \\
                          & \makecell[r]{\textbf{MR}}  &\textbf{MRR}  &\textbf{Hits@10/3/1} 
                          & \makecell[r]{\textbf{MR}} &\textbf{MRR}  & \textbf{Hits@10/3/1}\\ 
 
  \hline
  TransE \citep{transE}&\makecell[r]{39,140}&0.000&\ \ 0.0/\ \ 0.0/\ \ 0.0&\makecell[r]{12,311}&0.025&\ \ 5.8/\ \ 2.1/\ \ 0.7\\
  TuckER \citep{TuckER}&\makecell[r]{152,457}&0.000&\ \ 0.0/\ \ 0.0/\ \ 0.0&\makecell[r]{3,401}&0.285&50.5/35.3/20.1\\
  \hline
  TEE (CNN+TransE) &\makecell[r]{31,161}&0.004&\ \ 3.6/\ \ 1.3/\ \ 0.4&\makecell[r]{4,754}&0.013&16.7/\ \ 7.9/\ \ 1.3\\
  TEE (BiLSTM+TransE) &\makecell[r]{18,005}&0.009&\ \ 6.5/\ \ 2.5/\ \ 0.9&\makecell[r]{6,271}&0.011&15.4/\ \ 6.7/\ \ 1.1\\
  TEE (CNN+TuckER) &\makecell[r]{3,342}&0.181&23.5/19.9/14.5&\makecell[r]{\textbf{1,355}}&\textbf{0.553}&\textbf{74.5/62.7/45.3}\\
  TEE (BiLSTM+TuckER) &\makecell[r]{\textbf{1,562}}&\textbf{0.461}&\textbf{55.7/47.4/41.4}&\makecell[r]{1,702}&0.113&24.0/12.4/\ \ 5.9\\
  \hline
  \end{tabular}
  \vspace{-1mm}
  \caption{Experimental results of TuckER and our models on ATOMIC and ConceptNet-100k}
  \label{experiments}
\vspace{-5mm}
\end{table*}

Since the \textit{TuckER} model in the original paper \citep{TuckER} train entire graphs at each training batch and the CKGs contain a much larger amount of entities than normal KGs, it leads to intolerable memory cost when we apply the training method directly on CKGs. We revise the training function by taking sampling methods used by \citet{transE, TransR}, using one positive entity and one random negative entity per training case. We call the TuckER Model with the original training method as \textit{original TuckER} and ours as \textit{revised TuckER}. The training objective of the original TuckER is 
\begin{equation}
  L = \frac{1}{n_e}\sum_{i=1}^{n_e}({\bf y}^{(i)}log({\bf p}^{i})+({1- \bf y}^{(i)})log({1-\bf p}^{(i)}))
\end{equation}
where $n_e$ is all entities, $\bf p \in \mathbb{R}^{n_e}$ is the vector of predicted probabilities and $\bf y \in \mathbb{R}^{n_e}$  is the binary label vector. 
For our revised TuckER, the training objective is the same, but 
$n_e$ denotes the current training entities instead, 
$\bf p $ becomes $\in \mathbb{R}^{2}$, 
and $\bf y $ becomes $ \in \mathbb{R}^{2}$. 
To make fair in the comparison, the revised TuckER, instead of the original one, is used as the baseline model.

\section{Results}
\vspace{-1mm}

The results of the baseline models and our models are shown in Table \ref{experiments}.

\textbf{ATOMIC}.
Since no source entities of the testset in ATOMIC have trained embeddings by standard KGE models, the test tuples cannot be scored by the baseline models. As a result, the results of baselines on ATOMIC are equal to random guesses. 
TEE (CNN+TransE), TEE (BiLSTM+TransE), and TEE (CNN+TuckER), TEE (BiLSTM+TuckER) have significantly better results over the baselines, which showing the ability of \textbf{Text Encoder Enhanced} model in handling \textit{unseen entities}. 
We find that 86\% of the words in the ATOMIC test entities appear in the training entities, which can explain why the TEE model can solve the \textit{unseen-entities problem} on ATOMIC. 

\textbf{ConcoptNet-100k}.
Both TEE (CNN+TransE) and TEE (BiLSTM+ TransE) perform better in Mean Rank and Hits@10/3/1, but worse in MRR.
TEE (CNN+TuckER) has better performance in all metrics compared with the baseline TuckER model. The results show that even though most test source entities are \textit{seen} at training data, adding text information can also benefit the results. 
However, TEE (BiLSTM + TuckER) has a negative performance in most metrics on ConceptNet-100k compared with TuckER, which indicates that if the entities are too short and the baseline model already performs well, using BiLSTM to encode them may cause overfitting.

TEE (BiLSTM + TuckER) performs better than TEE (CNN + TuckER) on ATOMIC but worse on ConceptNet-100k. This is because the entities of ATOMIC have longer text than entities of ConceptNet-100k, which is 4.41 to 1.72. BiLSTM may be stronger than CNN in encoding longer text. However, when encoding short text with one or two words, BiLSTM underperforms CNN.

\vspace{-1mm}
\section{Related Work}


\textbf{Standard KGE Models.}
A range of models have investigated on how to represent a knowledge graph, including TransE \citep{transE}, NTN \citep{NTN}, TuckER \citep{TuckER}, etc.
Standard KGE models cannot work in ATOMIC because the source entities in the testing set do not appear in the training set. 
External Text Enhanced KGE Models, such as DKRL \citep{DKRL} and Jointly \citep{Jointly}, are also related to our work. They use descriptions of entities crawled from Wikipedia to enhance the entity representation,  but do not encode the entity text information. In contrast, we concentrate on the entity text information yet do not import external descriptions of entities.


\textbf{\citet{malaviya2019exploiting}} is a concurrent paper that considers link prediction task on ConceptNet-100k and ATOMIC.
However, there are some key differences. 
First, the goal is different. While they try to apply off-shelf KGE models for CKGs completion, testing their performance, our goal is to do CKGR. 
Second, it does not solve the \textit{unseen-entities problem} because \textit{unseen entities} cannot get proper embedding in their testing period. Their ATOMIC results are reported using a different partition from existing work, which does not contain \textit{unseen entities}. In addition, our model can train over the full graph in a reasonable time/memory cost and do not need \textit{``subgraph sampling"}, which theirs need.

\vspace{-1mm}
\section{Conclusion}
\vspace{-1mm}

We compared the advantages of two types of methods for commonsense knowledge graph reasoning, namely, the generation method and the selection method. The generation method requires large human labor to evaluate, and we empirically find that it does not actually generate new semantic entities. 
The selection method faces \textit{rich textual information problem} and \textit{unseen-entities problem}. We solve the issues by combining the structure of neural Text Encoder and Knowledge Graph Embedding models. Experiments show that our model can outperform standard KGE baselines.
To our knowledge, we are the first to build a selection method for CKGR that allows unseen entities in the test data.

\bibliographystyle{acl_natbib}
\bibliography{anthology,emnlp2020}

\appendix
\label{sec:appendix}

\begin{table*}
  \centering
  \begin{tabular}{c|ccccccc}
  \hline

  {\textbf{Dataset}} & \makecell[c]{Tuple Number\\train/valid/test} &\makecell[c]{Entity \\Number} &  \makecell[c]{Relation\\ Number}&  \makecell[c]{Word \\Number}&  \makecell[c]{Average\\ Length}& \makecell[c]{\textit{Unseen} Tuple \\Proportion}\\ 
  \hline
  ConceptNet-100K &10000/1200/1200& 78279 & 34 & 21498 & 1.72 & 2.8\%\\
  ATOMIC & 709996/79600/87481 &304903 & 9 & 29415 & 4.41 & 100\% \\
  \hline
  \end{tabular}
  \vspace{-2mm}
  \caption{\label{dataset-details}
  Detailed data of the two datasets. The first/second/third column is the tuple/entity/relation number of the dataset; The forth column is the word number of entities and relations; The next column is the average length of entities' texts. The last column is the proportion of unseen test tuples in all test tuples.
  }
  \vspace{-3mm}
\end{table*}

\begin{table*}
  \centering
  \begin{tabular}{c|c|c|c}
  \hline
  
  {Input Source Entity} & \makecell[c]{Input Relation} &\makecell[c]{Output Target Entity \\ (Genrated by COMET)} &  \makecell[c]{Similar/Better Entity \\ in Training Set}\\ 
  \hline
go to get haircut&causes&hair to be cut&hair be become short
\\
man&desires&be with woman&have sex with girl
\\
telephone&is a&call call&communication tool
\\
food&is used for&keep you alive&stay alive
\\
read newspaper&motivated by goal&know about current event&learn about current event
\\
get up early&has a prequisite&be up early&wake up early
\\
mechanic&is capable of&repair car&repair car body
\\
mother&is capable of&love her child&mother her child
\\
pipe&is used for&hold tobacco&hold tobacco in pipe
\\
water&is capable of&be freeze water&be freeze
\\
needle&is used for&inject drug&\makecell[c]{inject substance into\\ someone body}
\\
have asthma attack&has a prequisite&have asthma&person with asthma
\\
car&receives action&park in garage&put car in garage
\\
submarine&is a&dangerous occupation&be dangerous occupation
\\
awake&requires&you feel groggy&feel groggy
\\
have haircut&causes&hair to grow&hair be become short
\\
music stand&is used for&hold your instrument&hold music
\\
brake&is capable of&stop car&slow vehicle
\\
get job&requires&you get pay&get pay
\\
triangle&has a&four side&3 side
\\
lake&is a&large lake&large body of water
\\
urinate&motivated by goal&you be urinate&\makecell[c]{eliminate excess fluid \\from your body}
\\
pillow case&is used for&keep pillow&keep pillow clean
\\
cat&is a&small than cat&carnivors
\\
there&is a&no way to predict future&no proof of god
\\
eat breakfast in bed&requires&spill coffee on sheet&spill coffee on bedcover
\\
screwdriver&is capable of&fit screw&tighten screw
\\
high rise&is a&call skyscraper&skyscraper
\\
magician&is capable of&trick person&perform trick
\\
surprise someone&has a prequisite&surprise them&not tell your plan to them
\\
horse&receives action&train to jump over obstacle&be train ot jump over obstacle
\\
chess board&has a&many side&4 side
\\
rain&is a&form of rain&weather phenomenon
\\
chat with friend&requires&talk about work&have something to talk about
\\
tickle&requires&tickle someone&find someone you can tickle
\\
take bus&has a prequisite&get bus pass&buy bus pass
\\
nurse&is capable of&clean your wound&clean wound
\\
terrorist&is capable of&attack other country&plan attack
\\
play game&has a prequisite&have game&have game to play
\\
\hline
  \end{tabular}
  \caption{\label{dataset-details}
  Detailed data of ConceptNet-100k related to (b) of Table 2. The first column is the Source Entity; the second is the Relation; the third is Target Entity generated by COMET; the last column is the corresponding similar/better entity in the training data to the generated entity. (Some (source entity + relation)s repeats, we combine them together, so there are less than 43 lines.)
  }
  \vspace{-5mm}
  \end{table*}

\begin{table*}
  \centering
  \begin{tabular}{c|c|c|c}
  \hline
  
{Input Source Entity} & \makecell[c]{Input \\ Relation} &\makecell[c]{Output Target Entity \\ (Genrated by COMET)} &  \makecell[c]{Similar/Better Entity \\in Training Set}\\ 

\hline
PersonX likes the movie&xEffect&buys the ticket&He buys the tickets.
\\
PersonX compares \_\_\_ with another&xIntent&to compare&to compare the two
\\
PersonX walks in front of PersonY&xIntent&to be in front&to be in front of personY
\\
PersonX transmits in PersonY form&oWant&to get personx’s form back&to get it back
\\
PersonX reports every \_\_\_&xNeed&to have a report&to have a report of the facts
\\
PersonX compares \_\_\_ with another&xNeed&to have a different opinion&has different opinions
\\
PersonX gives it PersonX’s name&xNeed&to have a name&to have name
\\
PersonX puts PersonY’s head down&xNeed&to grab persony’s head&to grab PersonY’s head
\\
PersonX takes down a peg&xNeed&to have a peg&to get a peg
\\
PersonX transmits in PersonY form&xNeed&to have a form&get a form
\\
PersonX wants to learn to swim&xNeed&to have a swimming suit&get a swimming suit
\\
PersonX tells my mother&xNeed&to call my mother&to call my mom
\\
PersonX grabs a towel&xNeed&to find a towel&to find a towel to dry off.
\\
PersonX takes the offer&xNeed&to have a proposal&to make a proposal
\\
PersonX takes the offer&xNeed&to have a proposal&to make a proposal
\\
PersonX finds PersonX’s daughter&xNeed&to search for their daughter&to go look for her
\\
PersonX creates PersonY structure&xNeed&to know how to construct&to learn how to construct it
\\
PersonX visits PersonY’s cousin&xNeed&\makecell[c]{to drive to persony’s \\cousin’s house}&to drive to his place
\\
PersonX wants to learn to swim&xNeed&to have a swimming suit&get a swimming suit
\\
PersonX keeps \_\_\_ with the growth&xWant&to keep the growth going&\makecell[c]{to increase the growth\\ to next level}
\\
PersonX compares \_\_\_ with another&xWant&to compare other things&to compare the two
\\
PersonX bosses PersonY around&xWant&to greet persony’s wife&to greet PersonY’s wife
\\
PersonX bosses PersonY around&xWant&\makecell[c]{to get persony\\ to do their work}&\makecell[c]{to get PersonY to do \\what PersonX wants}
\\
PersonX pulls PersonY’s hand away&xWant&to let go of persony’s hand&to let go of PersonY’s hand
\\
PersonX transmits in PersonY form&xWant&to get persony’s attention&to get PersonY’s attention
\\
PersonX throws food&xWant&to throw more food&to throw food away
\\
PersonX tells my mother&xWant&to tell my mother&to talk to my mom
\\
PersonX never been to one&xWant&to go to the one&to go to it
\\
PersonX sees \_\_\_ online&xWant&to read the reviews&to read reviews
\\
PersonX takes \_\_\_ up on the offer&xWant&to make a good offer&To make X a good offer
\\
PersonX finds PersonX’s daughter&xWant&to hug daughter&to hug their daughter
\\
PersonX visits PersonY’s cousin&xWant&to talk to persony’s cousin&to chat with the cousin
\\

\hline
  \end{tabular}
  \caption{\label{dataset-details}
  Detailed data of ATOMIC related to (b) of Table 2. The first column is the Source Entity; the second is the Relation; the third is Target Entity generated by COMET; the last column is the corresponding similar/better entity in the training data to the generated entity.
  }
  \vspace{-5mm}
\end{table*}

\end{document}